\documentclass[conference]{IEEEtran}

\IEEEoverridecommandlockouts
\usepackage{cite}
\usepackage{amsmath,amssymb,amsfonts}
\usepackage{algorithmic}
\usepackage{graphicx}
\usepackage{textcomp}
\usepackage{xcolor}
\usepackage[]{footmisc}
\usepackage{cuted}
\usepackage{amsmath}

\usepackage[ruled,vlined]{algorithm2e}
\usepackage{listings}
\usepackage{hyperref}
\def\BibTeX{{\rm B\kern-.05em{\sc i\kern-.025em b}\kern-.08em
    T\kern-.1667em\lower.7ex\hbox{E}\kern-.125emX}}
\begin{document}

\title{IGN :  Improved Training of Implicit Network with Generative method }

\author{\IEEEauthorblockN{Haozheng Luo, Tianyi Wu, Colin Feiyu Han, Zhijun Yan}

\IEEEauthorblockA{\textit{Computer Science} \\
\textit{Northwestern University}\\
\{robinluo2022,tianyiwu2023,colinfhan, ZhijunYan2022\}@u.northwestern.edu}
}
\maketitle

\begin{abstract}
In this work, we build recent advances in distributional reinforcement learning to give a state-of-art distributional variant of the model based on the IQN. We achieve this by using the GAN model's generator and discriminator function with the quantile regression to approximate the full quantile value for the state-action return distribution. We demonstrate improved performance on our baseline dataset - 57 Atari 2600 games in the ALE. Also, we use our algorithm to show the state-of-art training performance of risk-sensitive policies in Atari games with the policy optimization and evaluation.
\end{abstract}

\begin{IEEEkeywords}
Reinforcement Learning, Implicit Quantile Networks, Generative adversarial network
\end{IEEEkeywords}

\section{Introduction}
Distributional reinforcement learning \cite{sobel_1982,zhou2024optimizing} focus on the reward with intrinsic randomness within the reinforcement learning framework. Distributional reinforcement learning aims to model the distribution over returns and to use these distributions to evaluate and optimize a policy. Any distributional reinforcement learning algorithm is characterized by two aspects: the parameter of the return distribution and
the distance metric or loss function being optimized. Distributional reinforcement learning with Quantile regression \cite{dabney2017distributional} gives a considerable direction for Distributional reinforcement learning. The work based on the Deep Q-network \cite{mnih2013playing} became a popular direction in the reinforcement learning research work, especially in the game theory. 
It helps the distribution over returns is modelled explicitly instead of only estimating the mean. In recent years, a lot of work shows a good improvement in the game theory with the reinforcement learning policy, such as  Implicit Quantile Networks (IQN) \cite{dabney2018implicit} and Fully Parameterized Quantile Function (FQF) \cite{yang2020fully}. They give a great improvement on the policy, especially risk-sensitive policy. However, those works still show limited work. How to use the Generative Network to help improve the performance of the Distributional reinforcement learning has become a controversial problem. Because our training resource is 57 Atari 2600 games in the ALE, which is high-quality images, it gives us an idea using the transfer learning to improve the performance of the algorithm in our work. We introduced a new model called Implicit Generative Networks (IGN) based on the IQN. The Generative Network can work by sending the embeddings of the source and target task to the discriminator and using the generator to generate the learning thread of the model. The resulting loss is then (inversely) backpropagated through the encoder. This method is also regarded as one kind of transfer learning \cite{fregier2021mind2mind} which can significantly reduce training time. In our work, we freeze both the critic and generator of the original GAN's low-level layers. Also, we designed a method that gives out an auto-encoder constraint in order to compatible the internal representations of the critic and the generator. This assumption explains the accelerate in training time. Also, the generator can provide us with a clear boundary of coverage to help the model acceleration the training. In our experiment, we will use 57 Atari 2600 games in the ALE as the dataset to compare with our baselines, such as IQN and FQF, to show start-of-art performance on policy. The main contributions are as follows:

\begin{itemize}
    \item Using the generator and discriminator to rewrite the quantile function in the IQN model
    \item Build our new algorithm IGN with the base of the IQN model
    \item Show our algorithm state-of-art performance with the baselines in the Atari dataset.
\end{itemize}

\section{Related Work}
\subsection{Dataset}
\begin{figure}[htp]
\centering
\includegraphics[width=8cm]{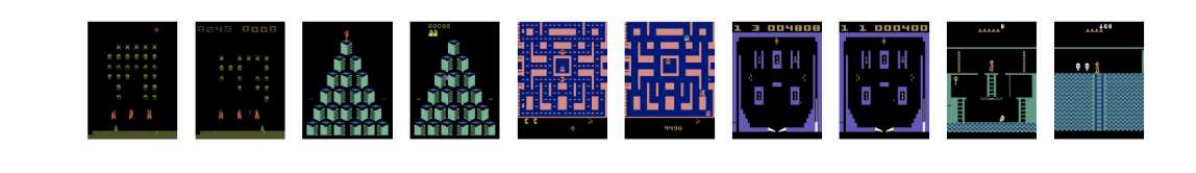}
\caption{Atari Game}
\label{fig:Atari}
\end{figure}

There are a lot of datasets we can use to show our models' performance and compare them. In our work, we want to compare our model with baselines with several classic datasets. One of them is Atari games datasets (see Fig.~\ref{fig:Atari}) \cite{kurin2017atari}, which is one of the classic datasets in reinforcement learning in game theory. We use the 57 Atari 2600 games in the ALE, and it can help show our contribution to our algorithm in the risk-sensitive policies, in which the Atari 2600 games have ~200k frames collected using the OpenAI baseline and implemented based on Advantage Actor Critic (A2C). Also, they use sticky actions to make the problem become more challenging. There is a $25\%$ probability that the agent’s previous action is executed. It will greatly help us show our algorithm performance on risk-sensitive policy.

\subsection{Quantile Networks and Transfer Learning}
Before our work, there was a lot of research already work in reinforcement learning. Implicit Quantile Networks (IQN) \cite{dabney2018implicit} is one of the models that shows a good result. It brings a huge improvement on the risk-sensitive policy training. However, the model based on the Quantile distribution make a huge fragile on the performance of the model, and it should cost huge time for training the model. We expected to design a state-of-art algorithm that has more robustness and has more effectiveness in training the policy. One of the possible methods is using transfer learning. Transfer learning \cite{zhuang2020comprehensive} is an optimization method that can improve the performance of the model or rapid progress in the second task. In our research, transfer learning can help us have a higher start and higher slope from the second training. Also, it will have great help in the policy evaluation. There are some state-of-art transfer learning methods, such as VGG \cite{simonyan2015deep}, ResNet \cite{he2015deep} Model. From several model pooling, there is one algorithm using Gradient method is GAN model, which can help our model have a more competitive algorithm with origin IQN model

\subsection{Existing Offline Policy Evaluation Methods}
Most existing offline policy evaluation methods under the above setting with mean criteria can be grouped into three categories. The first category is direct methods, which directly estimate the state-action value function \cite{luckett2019estimating,shi2020statistical}, also known as the Q-function. The second category of offline policy evaluation methods is motivated by marginal importance sampling \cite{liu2018breaking}. The last category of methods combines direct and marginal importance sampling methods to construct some doubly robust estimator, which is also motivated by the efficient influence function like equation  \ref{eq1} \cite{kosorok2019precision}. In which,  $Q^{\pi}$ and $\omega^{\pi}$ are two nuisance functions.

\begin{multline}
    \frac{1}{T}\sum_{t=0}^{T-1}\omega (S_t,A_t) (R_t + \\ \gamma \sum_{a^\prime \in \mathcal{A}} \pi(a^\prime |S_{t+1})Q^{\pi} (S_{t+1},a^{\prime} ) - Q^{\pi} (S_t,A_t)) \\+ (1 - \gamma) \mathbb{E}_{S_0 \sim \mathbb{G}} [\sum_{a_0 \in \mathcal{A} } \pi(a_0|S_0)Q^{\pi} (S_0, a_0)] - \mathcal{V}(\pi)
\label{eq1}
\end{multline}

In which, the $V(\pi)$ has the formula as eq \ref{eq2}.
\begin{equation}
\label{eq2}
    \mathcal{V}(\pi) = (1 - \gamma)\int_{s \in \mathcal{S}} \sum_{a \in \mathcal{A}} \pi(a|s) Q^{pi} (s, a) \mathbb{G}(ds)
\end{equation}

\subsection{Gradient Method}
In our work, we hope to use the generative method to improve our model. The most popular method on the generative method is Generative Adversarial Networks (GAN) \cite{goodfellow2014generative}. The GAN model shows a great improvement for our model in the gradient periods and make our model have high efficiency to get the best reward score. 

\begin{figure}[htp]
\centering
\includegraphics[width=8cm]{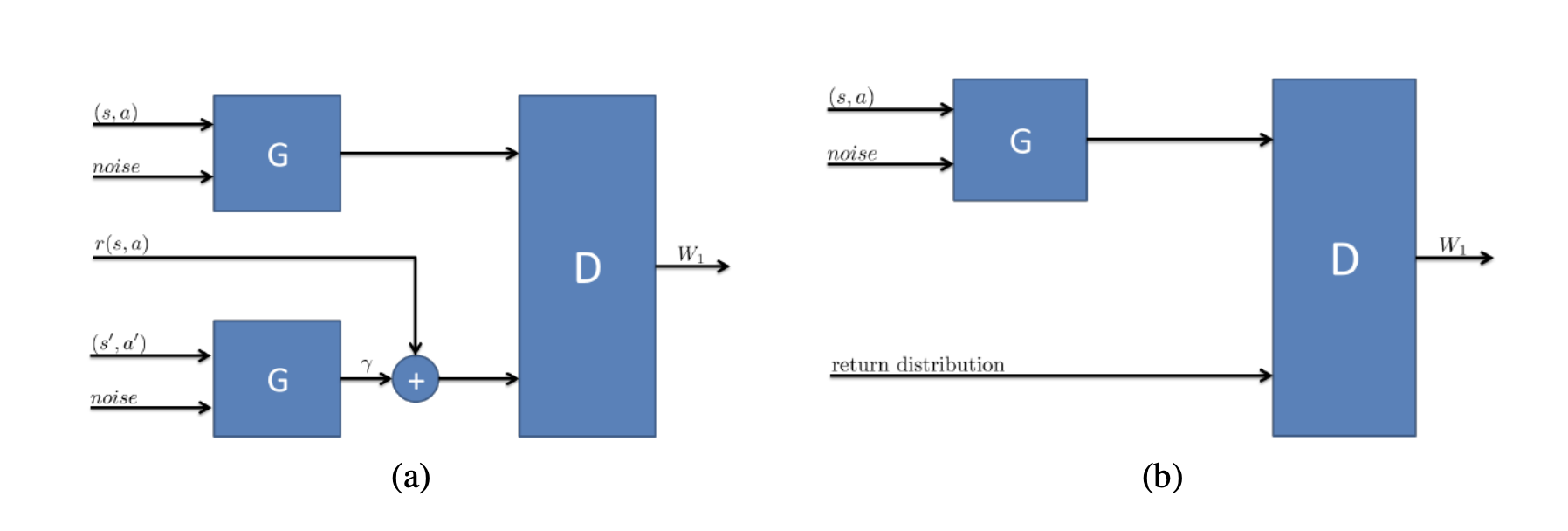}
\caption{GAN configurations.  \cite{freirich2019distributional} (a) Bellman-GAN; (b) WGAN \cite{gulrajani2017improved}}
\label{fig:gan}
\end{figure}

GANs can train two competing models (typically NNs) \cite{goodfellow2014generative}. The \textit{generator} takes noise $z \sim P_z$ as input and generates samples according to some transformation, $G_\theta(z)$. The \textit{discriminator} takes samples from both the generator output and the training set as input and aims to distinguish between the input sources. Goodfellow \cite{goodfellow2014generative} measured the discrepancy between the generated and the real distribution using the Kullback-Leibler divergence. However, this approach was improved by Arjovsky \cite{gulrajani2017improved} by using the Wasserstein-1 distance (as eq \ref{eq3}) \cite{freirich2019distributional}. WassersteinGANs exploit the Kantorovich-Rubinstein duality \cite{villani2009optimal}, 

\begin{equation}
\label{eq3}
    W_1(\mathbb{P}_r,\mathbb{P}_g)=\sup_{f\in1-Lip}\{\mathbb{E}_{x\sim\mathbb{P}_r}f(x)-\mathbb{E}_{x\sim\mathbb{P}_g}f(x)\}
\end{equation}
where $1-\text{Lip}$ is the class of Lipschitz functions with Lipschitz constant 1, in order to approximate the distance between the real distribution, $\mathbb{P}_r$, and the generated one, $\mathbb{P}_g$. The GAN objective is then to train a generator model $G_\theta(z)$ with noise distribution $P_z$ at its input, and a critic $f \sim 1- \text{Lip}$,  achieving
\begin{equation}
\label{wgan-gp}
    \min_{G_\theta}\max_{f \sim 1-{Lip}}\{\mathbb{E}_{z\sim\mathbb{P}_r}f(x)-\mathbb{E}_{z\sim\mathbb{P}_z}f(G_\theta(z))\}
\end{equation}
The WGAN-GP\cite{gulrajani2017improved} (Fig.~\ref{fig:gan}(b)) is a stable training algorithm for equation \ref{wgan-gp}  \cite{freirich2019distributional} that employs stochastic gradient descent and a penalty on the norm of the gradient of the critic with respect to its input.

\section{Assumption}
\label{app:a}
AS1: Markov assumption with stationary transitions: there exists a transition function $p$ such that
\begin{equation}
 Pr(S_{t+1} = s^{\prime}|A_t = a, S_t = s, (S_j ;A_j ;R_j)_{0 \geq j < t}) = p(s^{\prime}|a,s);   
\end{equation}
for any $t \geq  0, a \in \mathop{A}$ and $s, s^{\prime} \in \mathop{S}.$

AS2: $R_t$ is a known function of $S_t$, $A_t$ and $S_{t+1}$, i.e, $R_t =\mathop{R}(S_t,A_t, S_{t+1})$, for any $t \geq 0$. We
further assume the range of $\mathop{R}$ is $[- R_{max},R_{max}]$.

AS3: (Behavior policy) the data are generated by axed stationary policy b, which may
not be known. We usually call b behavior policy.

\section{Method}
\subsection{Task Definition}
The problem is driven by recent development in distributional reinforcement learning. \cite{dabney2018implicit} Consider a single trajectory ${(S_t, A_t, R_t)}_t  \geq 0$ where $(S_t, A_t, R_t)$ denotes the state-action-reward triplet collected at time t. We use $S 	\subseteq \mathbb{R}_p$ and $\mathcal{A}$ denote the state and action space, respectively. We assume A is discrete and finite, and rewards $R_t$ are uniformly bounded by $R_{max}$. Consider the offline setting. Then we observed data consisting of N trajectories, corresponding to N independent and identically distributed copies of ${(S_t, A_t, R_t)}_t  \geq 0$. For any i = $1,\cdots ,N,$ data collected from the $i^{th}$ trajectory can be summarized by ${(S_{i,t}, A_{i,t}, R_{i,t}, S_{i,t+1})}0 \leq t < T_i$ , where $T_i$ denotes the termination time. For simplicity, we assume $T_i$ = T for i = $1, \cdots ,N.$

A policy defines the agent’s way of choosing the action at each decision time. We focus on the stationary policy. Specifically, at each time point t, a stationary policy $\pi$ maps the current state value into a probability mass function over the action space, i.e., $\pi(a|s)$ denotes the probability of choosing action a given the state value s.

In our work, we will work on the policy evaluation and policy optimization. In policy evaluation, we want to see the performance of the model what would have happened in the absence of a policy. We will use the Pong environment agent as a fixed agent, which we will use as one of the offline policy, and we will use our model to train the online policy and see the performance of the model training. In our work, we will compare the reward scores and efficiency of the timestamps because comparing with the complex model with a simple model in the resource using is not a fit game, we will compare the timestamps( measured with the training cycle time, not the real time). In our future work, if there is one model that requires less real time is one of the potential research directions in our work. For the policy optimization, we will train a model with different environments to evaluate our model performance in timestamps and reward gain with other baselines models.

\subsection{Distributional Reinforcement Learning}
We introduce a state that uses quantile regression to approximate risk-sensitive strategies, in other words, an optimization method for reinforcement learning via a full quantile function of the behavioural reward distribution.

We always model the agent and the environment as a Markov decision model\cite{von2007theory} in classical reinforcement learning designs. Here, we define the rule $\pi(\cdot|x)$ as a mapping from states to distributions constantly changing with actions. For any agent that follows the rule $\pi$, we define the discounted sum of future rewards as a function $Z$. Clearly, $Z^\pi (x,i)$ varies with state and action as following
\begin{equation}
    Z^\Pi(x,i)=\sum\limits_{t=0}^\infty \beta^t R(x_t, a_t)
\end{equation}
, where $\beta$ is a discount factor and $\beta \in (0,1)$. The action-value function can be defined as $Q^\pi(x,i)=\mathbb{E}[Z^\pi (x,i)]$, in which $\mathbb{E}$ is the Bellman equation operator\cite{bellman1957dynamic}.
\begin{equation}
    Q^\Pi(x,i)=\mathbb{E}[R(x,i)]+\beta\mathbb{E}_{P,\Pi}[Q^\Pi(x',i')]
\end{equation}
Azer's research\cite{azar2012sample} on reinforcement learning proposed a distributed reinforcement learning theory that values the benefit distribution $Z^\pi(x,i)$ rather than the action-value function $Q^\pi(x,i)$. According to the analysis\cite{barth2018distributed}, it can be found that distributed reinforcement learning methods have substantial advantages in terms of sample complexity, final performance, and hyperparameter processing robustness.

\subsection{Quantile Regression Approaching}
According to Bellemare's study\cite{bellemare2017distributional}, the distributed Bellman operator is a contraction in the p-Wasserstein metric. There are two main theories for minimizing the Wasserstein metric. One is distributed reinforcement learning using quantile regression and shows that the resulting predictive distributed Bellman operator is a contraction of the $\infty$-Wasserstein metric by properly selecting the quantile target\cite{dabney2018distributional}. Another study shows that the original class of classification algorithms is a contraction in the Cramér distance, the L2 metric on the cumulative distribution function.\cite{rowland2018analysis}

QR-DQN minimizes the Wasserstein distance distribution Bellman objective by estimating the quantile function at precisely chosen points. This estimation uses quantile regression, which has been shown to converge to the true quantile function value when minimized using stochastic approximation. In QR-DQN, the stochastic returns are approximated by a homogeneous mixture of N Diracs (as equation \ref{eq:eq1}),
\begin{equation}
    \label{eq:eq1}
    Z_\theta(x,i):=\frac{1}{N}\sum\limits^N_{i=1}a_{\theta_i(x,i)}
\end{equation}

in which, $\theta_i$ is assigned a fixed quantile target $\tau$. We can use Huber quantile regression loss to train this quantile target $\tau$ with threshold $\delta$.
\begin{equation}
\rho^\delta_\tau(a_{ij})=|\tau-\mathbb{I}\{a_{ij}<0\}|\frac{L_\delta(a_{ij})}{\delta}\text{, with}
\end{equation}
\begin{equation}
L_{\delta}(a_{ij}) =\left\{
\begin{aligned}
\frac{1}{2} a_{ij}^2 &  & |a_{ij}| \leq \delta\\
\delta(|a_{ij}| - \frac{1}{2}\delta) & & |a_{ij}| > \delta 
\end{aligned}
\right.
\label{huber}
\end{equation}

In practice, we use the IQN as the baseline and implement the gradient method with the GAN model and calculate the GAN score as the gradient score and calculate the new gradient with the Huber loss (see equation \ref{huber}) to get the reward score and use this new reward score into the agent. 

\begin{strip}
\begin{equation}
    f^\pi_Y\left(y,s,a\right) =
    \mathbb{E}
    \left[ \sum\limits_{a^{\prime}\in\mathcal{A}}\pi(a^{\prime}|S_{t+1}) f^\pi_Y(\frac{y-R_t}{\gamma},S_{t+1},a^{\prime})|S_t=s,A_t=a \right]
\label{eq4}
\end{equation} 
\end{strip}

\subsection{Distributional Bellman Equation}
Define a random variable $Y=\sum\limits_{t=0}^{+\infty}\gamma^{t}R_t$, whose support is \\ $\left[-\frac{R_{max}}{1-\gamma},\frac{R_{max}}{1-\gamma}\right]$ by assumption. Given a target policy $\pi$, our goal is to estimate the distribution of $Y$ given any state
$s \in \mathcal{S}$ and $a \in \mathcal{A}$. We use $\mathcal{U}$ to denote the product space of cumulative rewards, state and actions, i.e., $\mathcal{U}=\left[-\frac{R_{max}}{1-\gamma},\frac{R_{max}}{1-\gamma}\right]\times\mathcal{S}\times\mathcal{A}$. For any $y \in \mathbb{R}$, define either a conditional density
or a conditional probability mass function of $Y=y$ given a state-action pair $(s, a)$ as
\begin{equation}
     f^\pi_Y\left(y|S_0=s,A_0=a\right) \triangleq f^\pi_Y\left(y,s,a\right)
\end{equation}
  
where $f^\pi_Y$ indicates actions are selected according to $\pi$. In order to estimate $f^\pi_Y$ given any $(s, a)$, we have the distributional Bellman equation that under Assumptions (AS1) and (AS2), for any $t \geq 0$, $y \in \mathbb{R}$, $s \in \mathcal{S}$ and $a \in \mathcal{A}$, the equation \ref{eq4} holds.

Motivated by the distributional Bellman operator is $\pi$-contraction under a modified Wasserstein distance, and the recently developed generative adversarial networks (GANs), we propose a Wasserstein GAN-based distributional policy evaluation method to estimate $f^\pi_Y$ will discuss the benefits of using GANs. For any continuous function $h$ defined over $\mathbb{R}\times\mathcal{S}\times\mathcal{A}$, the following equation holds.
\begin{strip}
\begin{equation}
     \begin{aligned}
    \mathbb{E}\left[\frac{1}{T}\sum\limits_{t=0}^{T-1}\mathbb{E}^\pi\left\{h\left(\sum\limits_{t'=t}^{+\infty}\gamma^{t'-t}R_{t'},S_t,A_t\right)|S_t,A_t\right\}\right]
    =\gamma\mathbb{E}
    \left[\frac{1}{T}\sum\limits_{t=0}^{T-1}
    \mathbb{E}^\pi
    \left\{h\left(\gamma\sum\limits_{t'=t+1}^{+\infty}\gamma^{t'-t-1}R_{t'}+R_t,S_t,A_t\right)
    |S_{t+1},R_t,S_t,A_t\right\}
    \right]
    \end{aligned}
\label{eq:bell-gan}
\end{equation}
\end{strip}

Motivated by equation \ref{eq:bell-gan}, we propose to use generative adversarial networks to estimate $f^\pi_Y$ of $Y$ given state-action pair $(s, a) \in \mathcal{S} \times \mathcal{A}$. Specifically, we aim to learn a generator $\mathbb{G}^\pi$ that takes $(s, a)$ a uniform random variable $Z$, i.e.,$Z \sim$ uniform (0, 1), as the input, and the output is a pseduo sample $\tilde{Y}$ as one realization of target conditional distribution $f^\pi_Y$. This is also inspired by the conditional GANs introduced by Mirza and Osindero\cite{mirza2014conditional}. In typical GANs, the generator is trained by minimizing some divergence between the conditional distribution of $Y|s, a$ and $\tilde{Y}|s,a$. Alternatively, a generator $\mathbb{G}^\pi(Z,S,A)$ is trained such that the joint distribution of $(Y(S,A), S, A)$ is the same as $(\mathbb{G}^\pi(Z,S,A), S, A)$, where $Y(S,A)$ is a random variable that has the same distribution of $Y$ given $s\in\mathcal{S}$ and $a\in\mathcal{A}$. However, different from conventional distribution learning tasks, neither $Y$ nor $Y(S,A)$ is not observed in our problem. Nevertheless, the distributional Bellman equation and the equation\ref{eq:bell-gan} indicate us to obtain $\mathbb{G}^\pi$ by solving the following min-max optimization problem, which has the similar spirit of Wasserstein-GANs\cite{arjovsky2017wasserstein}.
\begin{strip}
\begin{equation*}
        \mathop{\mathrm{minimize}}\limits_{\mathbb{G}} 
        \mathop{\mathrm{sup}}\limits_{h:\parallel h \parallel _{\text{Lip}}\leq 1} \\ \left\{\gamma\mathbb{E}\left[\frac{1}{T}\sum\limits_{t=0}^{T-1}\sum\limits_{a\in\mathcal{A}}\pi(a|S_{t+1})h(\gamma\mathbb{G}(Z_t,S_{t+1},a)+R_t,S_t,A_t)\right]\mathbb{E} \left[\frac{1}{T} \\\sum\limits_{t=0}^{T-1}h(\mathbb{G(Z_\textit{t},S_\textit{t},A_\textit{t}),S_\textit{t},A_{\textit{t}})}\right]\right\}
\end{equation*}
\label{eq:mim}
\end{strip}
where $\{Z_t\}_{0 \leq t \leq (T-1)}$ are set to independently follow uniform(0,1), and $h$ is a function mapping from $\mathbb{R}^{p+2}$ to $\mathbb{R}$.

In practice, as in conventional GANs, we restrict $h$ to some set of neural networks $\mathcal{R}_D$ in addition, which can provide tractable computation. In addition, by powerful deep neural networks, we can handle potential high-dimensional state space $\mathcal{S}$. Similarly, we also use some set of neural networks $\mathcal{R}_G$ to approximate the generator $\mathbb{G}^\pi$.

\subsection{Estimation and Algorithm}
Our goal is to find a generator $\mathbb{G}^\pi(\bullet, S, A)$ to approximate the conditional distribution of $Y$ given $S$ and $A$. This is, in general, a very challenging problem, especially when $S$ is a high-dimensional state, e.g., the number of states in $\mathcal{S}$ is huge. Thus, suggested by Haas and Richter  \cite{haas2020statistical}, we impose some structure assumptions on the underlying true distribution. In particular, we assume the distribution of $(Y (S,A), S,A)$ lie in a $d_{\mathbb{G}}$-dimensional subspace with $d_{\mathbb{G}} < p+2$. Based on this consideration, we define the following function class for the generator.

Let $\mathcal{G}(d_{\mathbb{G}}, \beta, K)$ be the class of all measurable functions $\mathbb{G}:\mathbb{R}^{p+2}\rightarrow\mathbb{R}$ such that any component only depends on $d_{\mathbb{G}}$ arguments and lies in $C^\beta(\mathcal{U},K)$.

Given the offline dataset $\mathcal{D}_N=\{(S_{i,t},A_{i,t},R_{i,t},S_{i,t+1})\}_{ 0 \leq t \leq (T-1),1 \leq i \leq N}$, we use the empirical average to approximate the objective function and solve the following optimization to estimate the generator $\mathbb{G}^\pi$.
\begin{strip}
\begin{equation}
    \mathop{\mathrm{minimize}}\limits_{\mathbb{G\in\mathcal{R}(L_\mathbb{G},\textbf{p}_\mathbb{G},\text{s}_\mathbb{G})}}
    \mathop{\mathrm{sup}}\limits_{h\in\mathcal{R}(L_h,p_h,s_h),	\parallel h	\parallel_\text{Lip} \leq 1}
    \end{equation}
    \begin{equation}
    \left\{
        \gamma\frac{1}{NT}\sum_{i=1}^N\sum_{t=0}^{T-1}\sum_{a\in\mathcal{A}}
        \pi\left(a|S_{i,t+1}\right)h(\gamma\mathbb{G}\left(Z_{i,t},S_{i,t+1},a\right)
            +R_{i,t},S_{i,t},A_{i,t}) \frac{1}{NT}\sum_{i=1}^N\sum_{t=0}^{T-1}h\left(\mathbb{G}\left(\tilde{Z}_{i,t},S_{i,t},A_{i,t}\right),S_{i,t},A_{i,t}\right)
        \right\}
\end{equation}
\label{eq:th1}
\end{strip}

where $\left\{Z_{i,t}\right\}_{1 \leq i \leq N; 0 \leq t \leq (T-1)}$ and $\left\{\tilde{Z}_{i,t}\right\}_{1 \leq i \leq N; 0 \leq t \leq (T-1)}$ are two independent samples from uniform (0, 1). The optimal solution of the optimization above always exists because the inner maximization problem is Lipschitz with respect to $\mathbb{G}$ and $\mathbb{G}$ is Lipschitz with respect to all parameters in $\mathcal{R}(L_\mathbb{G},p_\mathbb{G},s_\mathbb{G})$ with bounded support. Similarly, the inner maximization problem also has an optimal solution. Stochastic gradient descent algorithms as the state-of-the-art can be implemented to solve the above optimization problem. We denote the final minimizer as by $\hat{\mathbb{G}}^\pi$, which can generate a sequence of pseudo samples $\{\tilde{Y}^\pi_m(s,a)\}_{1 \leq m \leq M}$ for some integer $M$ for approximating the distribution of $Y$ given state-action pair $(s, a)$ under the target policy $\pi$. This provides a flexible way for estimating interesting quantities related to the target policy $\pi$.

In all our examples, if $\hat{\mathbb{G}}^{\pi}$ is consistent and $M$ (or $K$) diverges to infinity, $\hat{\mathcal{V}}(\pi)$, $\hat{q}^{\pi}_\tau(Y)$ and $\widehat{Var}^{\pi} (Y )$ are all consistent.

\begin{algorithm}[h]
\SetAlgoLined
\textbf{Input:} A generator $\hat{\mathbb{G}}^\pi$ and $(s,a)$. \\
\textbf{for} \textit{m} = 1 to \textit{M}: \textbf{do} \\
\qquad(a) Sample $\textit{Z} \sim$ uniform (0,1) \\
\qquad(b) Compute the pseduo sample $\tilde{Y}_i(s,a)=\hat{\mathbb{G}}^\pi(Z,s,a)$. \\
\textbf{Output} $\{Y_i(s,a)\}_{1 \leq m \leq M}$.
\caption{Generate pseudo samples to approximate $d^{\pi_{old}}$}
\label{alg: pseudo}
\end{algorithm}

\section{Mathematical Prove }
\textbf{Proof of equation \ref{eq4}} By the idea of distributional Bellman equation, we can show that for
any $y \in \mathbb{R}$, $s \in \mathop{S}$ and $a \in \mathop{A}$,

$f_Y^{\pi} (y; s; a) = \mathbb{E}[f^{\pi}(R_0 +\gamma \sum\limits_{t=1}^{+\infty} \gamma^{t-1} R_t = y| R_0, S_0 = s, A_0 = a) | S_0 = s, A_0 = a]$

= $\mathbb{E}[f_{Y}^{\pi}(\sum\limits_{t=1}^{+\infty} \gamma^{t-1} R_t = \frac{y-R_0}{\gamma}| R_0, S_0 = s, A_0 = a) | S_0 = s, A_0 = a]$

= $\mathbb{E}[\sum\limits_{a^{\prime}\in \mathop{A}}\pi(a^{\prime}|S_1)
f_{Y}^{\pi}(\sum\limits_{t=1}^{+\infty} \gamma^{t-1} R_t = \frac{y-R_0}{\gamma}| S_1,A_1 = a^{\prime},R_0, S_0 = s;A_0 = a) | S_0 = s, A_0 = a]$

= $\mathbb{E}[\sum\limits_{a^{\prime}\in \mathop{A}}\pi(a^{\prime}|S_1)
f_{Y}^{\pi}(\frac{y-R}{\gamma},S_1, a^{\prime})| S_0 = s, A_0 = a]$

where the last inequality is due to Assumptions (AS1) and (AS2). By the stationary assumption in Assumptions (AS1) and (AS2), we can conclude that the above equation holds for any
$t \geq 0$.

\textbf{Proof of equation \ref{eq:bell-gan}} By Equation \ref{eq4}, one can obtain that
 $\mathbb{E}\left[\frac{1}{T}\sum\limits_{t=0}^{T-1}\mathbb{E}^\pi\left\{h\left(\sum\limits_{t'=t}^{+\infty}\gamma^{t'-t}R_{t'},S_t,A_t\right)|S_t,A_t\right\}\right]$
 
 = $\int_{S \times A}^{} \overrightarrow{d_{T}^b}(S,A)\int_{-\frac{R_{max}}{1-\gamma}}^{\frac{R_{max}}{1-\gamma}}h(y, a,s)f_Y^\pi(y, s, a)dydsda$
 
= $\int_{S \times A}^{}\int_{-\frac{R_{max}}{1-\gamma}}^{\frac{R_{max}}{1-\gamma}}\mathbb{E}[(\sum\limits_{a^{\prime} \in \mathop{A}}\pi(a^{\prime}|S^{\prime})f_Y^\pi(\frac{y-R}{\gamma}, S^{\prime}, a^{\prime})\\h(y, a,s))|S = s;A =a]\overrightarrow{d_{T}^b}(S,A)dsdadx$

=  $\gamma\int_{S \times A}^{}\int_{-\frac{R_{max}}{1-\gamma}}^{\frac{R_{max}}{1-\gamma}}\mathbb{E}[(\sum\limits_{a^{\prime} \in \mathop{A}}\pi(a^{\prime}|S^{\prime})f_Y^\pi(t, S^{\prime}, a^{\prime})\\h(\gamma t + R, a,s))|S = s;A =a]\overrightarrow{d_{T}^b}(S,A)dsdadx$

=  $\gamma\mathbb{E}\\\left[\frac{1}{T}\sum\limits_{t=0}^{T-1}\mathbb{E}^\pi\left\{h\left(\gamma\sum\limits_{t'=t+1}^{+\infty}\gamma^{t'-t-1}R_{t^{\prime}} + R_{t} ,S_t,A_t\right)|S_{t+1}, R_t, S_t,A_t\right\}\right]$

where we use change of variable in the calculus by letting $t = \frac{y-R}{\gamma}$ in the second equality.

Note that $\frac{R_{\text{max}}/(1 - \gamma) - R}{\gamma} \geq \frac{R_{\text{max}}/(1 - \gamma) - R_{\text{max}}}{\gamma} = \frac{R_{\text{max}}}{1 - \gamma}$, 
and $-\frac{R_{\text{max}}/(1 - \gamma) - R}{\gamma} \leq -\frac{R_{\text{max}}/(1 - \gamma) - R_{\text{max}}}{\gamma} = -\frac{R_{\text{max}}}{1 - \gamma}$

which implies the support of $t$ is the same as $Y$ . Therefore we conclude our proof.

\section{Theorem}
Assumption 1 (MDP structure). In addition to Assumptions (AS1)-(AS3), we further assume the state space $\mathop{S}$ and action space $\mathop{A}$ are discrete and finite. There exist a positive constant $p_{min}$ such that $\overrightarrow{d_T^{b}}(s,a)$ for every $s \in \mathop{S}$ and $a \in \mathop{A}$.

Assumption 2 (mixing Condition). The stochastic process $\{S_t,A_t,R_t\}_{t\geq0}$ is geometrically ergodic, i.e, there exists a function $\phi(s,a)$ and $\rho \in (0, 1)$ such that for any s, a.

where $||\bullet||_{TV} $indicates the total variation norm and $\mathbb{E}[\phi(S_0,A_0)] \leq C$ for some constant.

Define

Assumption 3: There exists a function $\mathbb{G}^{\pi} \in \mathcal{G}(d_{\mathbb{G}}; \beta,K)$ such that the distribution of 
$\mathbb{G} \in (Z, S,A)$ is the same as $(Y^\pi(S,A), S,A)$. For any $\mathbb{G} \in \mathcal{G}(d_{\mathbb{G}}, \beta, K) \cup  \mathop{R}(L_\mathbb{G}; p_\mathbb{G}; s_\mathbb{G})$,
there exists $\tilde{h} \in \mathop{C}^\theta(\mathcal{U}; \tilde{K}
)$, which only depends on $d_h$ arguments, for some positive constant
$\theta \geq 1$ and $K \in (0, 1)$ such that $\tilde{h} \in argmin_{h:||h||Lip\leq1}
\tilde{W}(h,\mathbb{G})$. Denote the class of $C^\theta(\mathcal{U}, \tilde{K})$
in which functions only depends on $d_h$ arguments as $\mathcal{H}(d_h; \beta; \tilde{K})$.
 
Assumption 4 (Neural Network). Define $\xi_{NT} = max\big\{(NT)^{-\frac{2\beta}{2\beta + d_{\mathbb{G}}}},(NT)^{-\frac{2\theta}{2\theta + d_{h}} }\big\}$. The
following conditions hold:

\begin{enumerate}
    \item $F > max(K, 1)$
    \item $s_{\mathbb{G}}  \asymp (NT)\xi_{NT}log(NT)$
    \item $\log_2(NT)\log_2(4 \max(d_{\mathbb{G}}, \beta, d_h, \theta)) \leq L_{\mathbb{G}} \lesssim \log(NT)$
    \item  $NT \xi_{NT} \leq \min \left\{ \min\limits_{1 \leq i \leq L_\mathbb{G}} P_{\mathbb{G}, i} NT \xi_{NT}, \min\limits_{1 \leq i \leq L_h} P_{h,i} \right\}$
    \item $L_f \leq L_{\mathbb{G}}; s_f \leq s_{\mathbb{G}}$
\end{enumerate}  

In this section, we establish some theoretical properties related to our method. Define a
random variable $Y^\pi(s, a)$ that has the same distribution of $Y$ given $(s, a)$ under the target policy $\pi$.

\section{Implementation}
Consider the neural network structure used by the IQN \cite{dabney2018implicit}. Let $\psi$ be the function of the functions computed by the convolutional layers and f be the function $\mathbb{R}^d \to \mathbb{R}^{|A|}$ with the subsequent fully-connected layers mapping $\psi(x)$ to the estimated action-values. Also, there is one additional function $\phi: [0, 1] → R^d$ computing an embedding for the sample point $\tau$.For our network, we use the same functions $\psi$, $\phi$ and f as in IQN. We make an improvement on the Quantile loss with our new methods.

Our training algorithm in IGN after the agent generates the states and actions, we use the original DQN model as the generator and use the three hidden layers neural network and ReLU activation function as the discriminator to train the model in critic with the algorithm \ref{alg: critic}  \cite{freirich2019distributional}. In the calculation of the Quantile loss, we put the result of the quantile result into the discriminator and use the equation \ref{eq: gan_loss} to get the GAN loss. Our IGN model optimizes with the gradient result of the GAN loss. Finally, we will use the Huber loss quantile result like an equation \ref{huber} to update the discriminator parameter in the IGN models .

Furthermore, We expected that N, the number of samples in  $\tau \sim$ U([0, 1]), would increase the sample complexity of IQN, with larger values leading to faster learning. Also, when we set the $\tau$ with N = 1, we would reach DQN performance. This would confirm the theory that many distributional RL algorithms' higher performance is due to their role as auxiliary loss functions, which would vanish if N = 1 was used.

To achieve model evaluation work, we make evaluation agent into our IQN model as the fixed policy. We use the fixed policy to get the result of the fixed network and use the new policy training in the online policy. We  use the same quantile loss calculation function in the policy evaluation. 

Also we use policy optimization and policy evaluation to show the performance of our IGN with the baselines. We implement the data collector function with Tensor-board logs. We also implement the Wasserstein-1 distance \cite{freirich2019distributional} and discriminator loss calculation function in the visualization method. This will be discussed more on the experiment part.

\begin{algorithm}[h]
\SetAlgoLined
 Train critic(repeat $n_critic$ times)\\
 (a) Sample $\{(s_t^{(i)}, a_t^{(i)}, r_t^{(i)}, s_{t+1}^{(i)}, a_{t+1}^{(i)})\}^{m}_{i=1}$ from replay pool\\
 (b) Sample both $\{z^{(i)}\}$ and $\{z^{\prime(i)}\}, i = 1,\cdots, m,$ from $P_{z}$ and $\epsilon^{(i)}, i = 1,\cdots, m,$ from U$[0,1]$ \\
 
 (c) $x_{\theta}^{(i)} = G_{\theta}(z^{i}|s_t^{(i)}, a_t^{(i)}),
 x_{\theta}^{\prime(i)} = r_{t}^{(i)} + \gamma G_{\theta}(z^{\prime(i)}|s_{t+1}^{(i)}, a_{t+1}^{(i)})$ 
\\
 (d) $\tilde{x}^{(i)} \gets \epsilon^{(i)} x_{\theta}^{(i)} + ( 1- \epsilon^{(i)})  x_{\theta}^{\prime(i)}$ \\
 (e) $g_{\omega} \gets \frac{1}{m} \bigtriangledown_{\omega} \sum_{i=1}^{m} [f_{\omega}(x_{\theta}^{(i)}) - f_{\omega}(x_{\theta}^{\prime(i)}) + \lambda(\parallel  \bigtriangledown_{\tilde{x}} f_{\omega}(\tilde{x}_{\theta}^{(i)}) 	\parallel-1)^2]$\\
 (f) $\omega \gets Adam(\omega, g_{\omega}, \alpha)$
 \caption{Train critic}
 \label{alg: critic}
\end{algorithm}

\begin{equation}
    g_\theta = - \frac{1}{m}  \bigtriangledown_{\theta} \sum_{i=1}^{m} [f_{\omega}(x_{\theta}^{(i)}) - f_{\omega}(x_{\theta}^{\prime(i)})]
\label{eq: gan_loss}
\end{equation}

\section{Experiment}
In order to show the state-of-art of our IGN model, we will show the performance of policy evaluation and policy optimization in our experiment. In our work, we compare IGN performance with the final reward and the timestamp the final reward get. And all of the work is based on the dataset - 57 Atari 2600 games in the ALE. We focus on two main questions, as follows.

\begin{enumerate}
    \item Whether the fixed model in IGN will use fewer timestamps in policy evaluation
    \item Whether our IGN model will use fewer timestamps to achieve the final reward than IQN, FQF. 
\end{enumerate}

Firstly, let us discuss policy optimization. We compare our method with IQN, FQF and other baselines from the efficiency of the model and best reward getting with the policy. And We measure Mean, Q-value and W-distance from eight different models (IQN, IGN, Seaquest, Qbert, Pong, Breakout, MsPacman, Freeway) and comparing with the result with our model.

Then in policy evaluation, we use several policies to train the specific environment with the fixed policy. In our work, we want to show whether the fixed policy can make new environment training become more efficient. We have Q-distribution for a specific evaluated policy based on different training steps (timestamps).

Also, we introduce the method of Monte Carlo Estimation to determine the policy. Monte Carlo Estimation is being used when state values are not enough since we do not have the model of the environment. The make use of this method will estimate action values instead of state values; for here, action values include LEFT, RIGHT, LEFTFIRE, RIGHTFIRE, NOOP and FIRE. The goal for this method is to have an estimation of the optimal action values. The result graphs show different actions with different scores. For comparison, Q-online estimation is also graphed for six actions.

\section{Result}
Firstly, we do the experiment to show the performance of our model work in the policy evaluation. The graphs on the left side will have Q-Fixed and Q-Online on the graph, where Q-Fixed is the Q-function of the fixed policy. The graphs on the right side will have Wasserstein-1 distance and Wasserstein-2 distance between Q-Fixed and Q-Online. The example graph is based on IQN and IGN for the Pong environment (Fig.~\ref{fig:fig4}, Fig.~\ref{fig:fig5}). For other fixed policies, graphs will be in the appendix. In our experiment, It is easy to realize our IGN can reduce the W-distance in the policy evaluation quicker than the IQN.

\begin{figure}[ht]
\centering
\includegraphics[width=7cm]{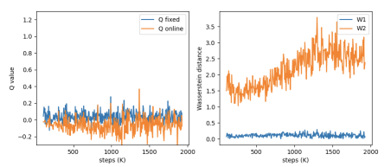}
\caption{Pong environment for Mean Q-value and W-distance graph based on IQN}
\label{fig:fig4}
\end{figure}

\begin{figure}[ht]
\centering
\includegraphics[width=7cm]{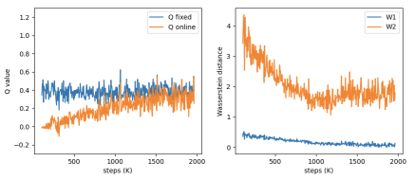}
\caption{Pong environment for Mean Q-value and W-distance graph based on IGN}
\label{fig:fig5}
\end{figure}

\begin{figure}[ht]
\centering
\includegraphics[width=5cm]{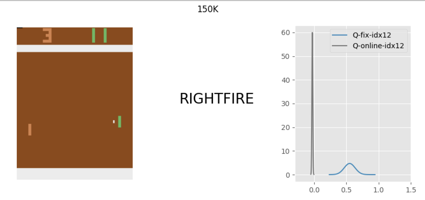}
\caption{ Q-distribution based on 150K training steps}
\label{fig:fig6}
\end{figure}

In order to see how the model evaluation work in our IGN model, we use the fixed policy ( Pong environment and IGN pretrained policy) to see how the training work in the several timestamps. In each graph, the left figure represents the state of the game; the top-mid number represents the training steps where k means thousands; the mid indicate current action; the right graph will show Q-distribution of Q-Fixed and Q-online where Q-value on the X-axis and density on the Y-axis. The example graph is being shown here for 150k training steps(Fig.~\ref{fig:fig6}) and 1700K training steps (Fig.~\ref{fig:fig7}), other results with different training steps will be in the appendix. 

\begin{figure}[ht]
\centering
\includegraphics[width=5cm]{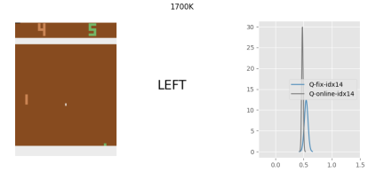}
\caption{Q-distribution based on 1700K training steps}
\label{fig:fig7}
\end{figure}
\begin{figure}[ht]
\centering
\includegraphics[width=5cm]{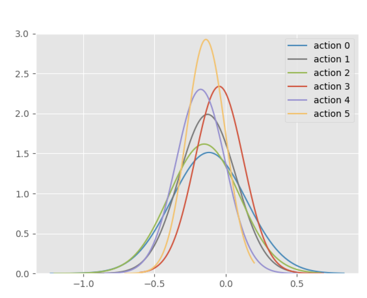}
\caption{Example graph for Monte Carlo Estimation on 6 actions}
\label{fig:fig8}
\end{figure}
\begin{figure}[ht]
\centering
\includegraphics[width=5cm]{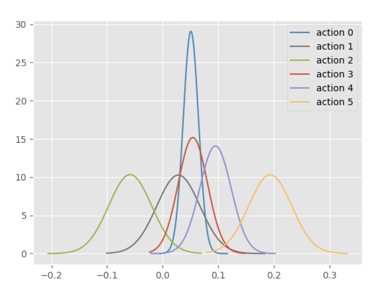}
\caption{Example graph for Q-online Estimation on 6 actions}
\label{fig:fig9}
\end{figure}

Also, we use the Monte Carlo Estimation and Q-online Estimation to visualize the estimation distribution of the reward in our policy evaluation work. The  Fig.~\ref{fig:fig8} shows Monte Carlo Estimation on six actions and the Fig.~\ref{fig:fig9} shows Q-online Estimation on six actions. The x-axis of both figures shows the reward of the actions getting in estimation. The y-axis shows the number of estimates getting a specific reward score for actions with a unit of thousands (k).

\begin{figure}[ht]
\centering
\includegraphics[width=7cm]{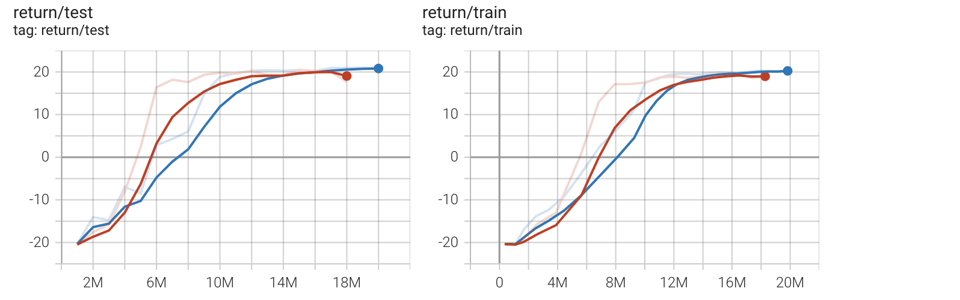}
\caption{Example graph for Policy Optimization}
\label{fig:fig10}
\end{figure}

Finally, we will use the tensor-board to show the result of the policy optimization result and compare it with other policy methods to show our model is states-of-art. The graphs will have the training process of IGN and IQN with blue and red color. The left graph shows the returned test result, and the right graph shows the returned train result. (Fig.~\ref{fig:fig10}). In the graph, we found our model IGN can get the same final reward score as the IQN model but use fewer timestamps to arrive at the highest reward score than the IQN model. It indicates our model can have efficiency in the model training. 

\section{Conclusion and Future work}
In our work, we make a new model, IGN, which shows a state-of-art performance in distributional reinforcement learning. To increase the algorithm's data efficiency, IGN may be trained using as little as one sample from each state-action value distribution or as many as computational restrictions allow. In addition, IGN enables us to broaden the scope of control rules to include a wide range of risk-sensitive policies linked to distortion risk measurements. Also, we show the state-of-art performance of our IGN with other baselines on the Atari-57 benchmark.
However, our model still has some limitations. Firstly, although our model needs less timestamp in the policy optimization and evaluation, the model still needs a lot of real-time training. Because we use the GAN model, it requires a higher cost than the IQN, FQF model. We hope we can find one method which has a similar or better performance than IGN but less cost required. Also, our model still has some sensitivity with the robustness of the noise in the training process work. We are considering using the transformer\cite{vaswani2017attention} and Kernel prepossessor \cite{07139} to see whether the model performance will become better.

\newpage
\bibliographystyle{IEEEtran}
\bibliography{sample-base}

\end{document}